\documentclass[11pt]{article}
\usepackage{scl}
\usepackage{times}
\usepackage{xurl}   
\usepackage{latexsym}

\usepackage{iftex}
\ifXeTeX
  \usepackage{fontspec, xunicode, xltxtra}
  \newcommand{\tibfontopts}{Script=Tibetan}
\else
  \usepackage{fontspec}
  \newcommand{\tibfontopts}{Renderer=HarfBuzz,Script=Tibetan}
\fi
\expandafter\newfontfamily\expandafter\tib\expandafter[\tibfontopts,Scale=0.9,Path=fonts/]{NotoSerifTibetan-Regular.otf}
\newfontfamily\zh[Scale=0.9,Path=fonts/]{NotoSerifTC-Regular.otf}

\usepackage{booktabs}
\usepackage{xcolor}
\usepackage{amsmath,xspace,ulem}
\usepackage{tikz}
\usetikzlibrary{positioning,arrows.meta}
\usepackage[hidelinks]{hyperref}   

\newcommand{\sa}[1]{\textit{#1}}
\newcommand{\tg}[1]{{\small\texttt{<#1>}}}
\newcommand{\tgc}[1]{{\small\texttt{</#1>}}}

\title{MITRA-lexicon: Lexica of attested translations\\ from Sanskrit into English, Tibetan and Chinese}

\author{Sebastian Nehrdich \\
  Center for Integrated \\ Japanese Studies \\
  Tohoku University \\
  Sendai, Japan \\
  {\small\texttt{nehrdich@tohoku.ac.jp}} \\\And
  Kengo Harimoto \\
  Center for Integrated \\ Japanese Studies \\
  Tohoku University \\
  Sendai, Japan \\
  {\small\texttt{kengo.harimoto.e6@tohoku.ac.jp}} \\\AND
  Sven Sellmer \\
  Institute of Oriental Studies \\
  Adam Mickiewicz University \\
  Pozna\'n, Poland \\
  {\small\texttt{sven@amu.edu.pl}} \\\And
  Oliver Hellwig \\
  ISLE \\
  University of Zurich \\
  Zurich, Switzerland \\
  {\small\texttt{oliver.hellwig@uzh.ch}} \\}

\date{}

\begin{document}
\maketitle
\begin{abstract}
The standard Sanskrit lexica are monumental and indispensable, but they cite their attestations as abbreviated references to editions, without links to the digitized texts now widely available and without a translation of the passage, and they naturally do not reflect translations made after them.
To address these gaps, we present MITRA-lexicon, a family of corpus-grounded lexica based on span annotations of translation equivalents with Sanskrit as the pivot: Sanskrit--English, derived from the Mitrasa\d{m}graha corpus, as well as Sanskrit--Tibetan and Sanskrit--Chinese, derived from the MITRA-parallel corpus of Buddhist literature. In total, the resource comprises 2.18 million headword entries in six directional lexica. 
Headwords are lemmas and compounds from silver annotations that follow the Digital Corpus of Sanskrit (DCS). Senses are ranked by frequency, and each of the 10.4 million attestations includes its sentence pair, with character offsets of the word and its equivalent. We propose a multi-step pipeline based on grammatical annotation models, fine-tuned large language models, and a teacher-student setup to produce the alignments. We evaluate the lexica against reference dictionaries and manually corrected gold alignments. We release the lexica and the fine-tuned span annotation model, together with a web interface for looking up headwords and their attested senses.
\end{abstract}

\section{Introduction}
\label{sec:intro}


The standard dictionaries of Sanskrit and of its Tibetan and Chinese translations give their textual evidence, where they give it at all, as abbreviated references to editions, without the translated passage and without information on frequency (Section~\ref{sec:related}).
What such dictionaries lack are attested translations: the renderings that translators actually chose for a word in specific passages, as opposed to glosses written by lexicographers, each tied to the sentence pair in which it occurs and counted over the whole corpus.
Beyond serving scholars who work with Sanskrit sources and their translations, a resource of attested translation equivalents with frequencies can support machine translation, which for Buddhist Chinese has been shown to benefit from retrieval augmentation \cite{nehrdich2025mitrazheval} and could use attested equivalents as terminology constraints.
It can also make AI-assisted applications in Buddhist Studies verifiable, as an interface that links each term of a machine translation to its lexicon entry lets users check every rendering against its attestations in human translations instead of taking the model's output on trust.

We build such resources automatically from parallel corpora of Sanskrit texts and their English, Tibetan and Chinese translations.
Every attestation of MITRA-lexicon is an aligned sentence pair with the span offsets of both the Sanskrit headword and its equivalent in another language, so that users can verify each entry against its source immediately (Figure~\ref{fig:alignment}).

\begin{figure}[t]
\begin{center}
\definecolor{alA}{HTML}{E69F00}
\definecolor{alB}{HTML}{56B4E9}
\definecolor{alC}{HTML}{009E73}
\definecolor{alD}{HTML}{D55E00}
\definecolor{alE}{HTML}{CC79A7}
\definecolor{alF}{HTML}{0072B2}
\tikzset{
  sw/.style={rounded corners=2pt, inner xsep=3.5pt, inner ysep=3pt, align=center,
             font=\small, minimum height=0.95cm},
  swal/.style={sw, fill=#1!28, draw=#1!80!black, line width=0.4pt},
  swno/.style={sw, text=black!55, draw=black!30, dashed, line width=0.4pt},
  tg/.style={inner xsep=1.5pt, inner ysep=2.5pt, font=\small, text depth=0.25ex, text height=1.6ex},
  tgal/.style={tg, rounded corners=2pt, fill=#1!28, draw=#1!80!black, line width=0.4pt},
  tgno/.style={tg, text=black!70},
  tib/.style={text height=2.1ex, text depth=1.3ex},
  lk/.style={draw=#1!85!black, line width=0.9pt, shorten >=1pt, shorten <=1pt, -{Stealth[length=4pt,width=4pt]}},
  rowlabel/.style={font=\scriptsize\sffamily, text=black!60, anchor=east},
}
\newcommand{\lem}[1]{\\[-1pt]{\scriptsize\itshape\color{black!60}#1}}
\begin{tikzpicture}[x=1cm,y=1cm]
\node[font=\small\bfseries, anchor=west] at (-1.9,1.55) {(a) Sanskrit--English};
\node[rowlabel] at (-0.35,0.95) {text};
\node[font=\small\itshape, anchor=west] at (-0.2,0.95) {hṛtāni ca mahīpānāṃ sarvaratnāni tejasā /};
\node[rowlabel] at (-0.35,0.03) {words};
\node[rowlabel] at (-0.35,-0.33) {lemmas};
\node[swal=alA, anchor=west] (a1) at (-0.2,-0.15) {hṛtāni\lem{hṛ}};
\node[swno,  right=5pt of a1] (a2) {ca\lem{ca}};
\node[swal=alB, right=5pt of a2] (a3) {mahīpānām\lem{mahīpa}};
\node[swal=alC, right=5pt of a3] (a4) {sarva-\lem{sarva}};
\node[swal=alD, right=1.5pt of a4] (a5) {ratnāni\lem{ratna}};
\node[swal=alE, right=5pt of a5] (a6) {tejasā\lem{tejas}};
\node[rowlabel] at (-0.35,-2.15) {translation};
\node[tgno, anchor=west] (e0) at (-0.2,-2.15) {He};
\node[tgal=alA, right=3pt of e0] (e1) {carried away};
\node[tgno, right=3pt of e1] (e2) {by};
\node[tgal=alE, right=3pt of e2] (e3) {force};
\node[tgal=alC, right=3pt of e3] (e4) {all};
\node[tgno, right=3pt of e4] (e5) {the};
\node[tgal=alD, right=3pt of e5] (e6) {jewels};
\node[tgno, right=3pt of e6] (e7) {of the};
\node[tgal=alB, right=3pt of e7] (e8) {kings};
\node[tgno, right=0pt of e8] (e9) {.};
\draw[lk=alA] (a1.south) to[out=-90,in=90] (e1.north);
\draw[lk=alB] (a3.south) to[out=-90,in=90] (e8.north);
\draw[lk=alC] (a4.south) to[out=-90,in=90] (e4.north);
\draw[lk=alD] (a5.south) to[out=-90,in=90] (e6.north);
\draw[lk=alE] (a6.south) to[out=-90,in=90] (e3.north);
\begin{scope}[yshift=-4.3cm]
\node[font=\small\bfseries, anchor=west] at (-1.9,0.85) {(b) Sanskrit--Tibetan};
\node[rowlabel] at (-0.35,0.03) {words};
\node[rowlabel] at (-0.35,-0.33) {lemmas};
\node[swal=alA, anchor=west] (b1) at (-0.2,-0.15) {vaśīkṛtāḥ\lem{vaśīkṛ}};
\node[swal=alB, right=5pt of b1] (b2) {sarva-\lem{sarva}};
\node[swal=alC, right=1.5pt of b2] (b3) {gaṇāḥ\lem{gaṇa}};
\node[swno,  right=5pt of b3] (b4) {sa\lem{tad}};
\node[swno,  right=5pt of b4] (b5) {vaḥ\lem{tvad}};
\node[swal=alE, right=5pt of b5] (b6) {svasti\lem{svasti}};
\node[swal=alF, right=5pt of b6] (b7) {kariṣyati\lem{kṛ}};
\node[rowlabel] at (-0.35,-2.15) {translation};
\node[tgal=alC, anchor=west, tib] (t1) at (-0.2,-2.15) {{\tib ཚོགས}};
\node[tgno, right=0pt of t1, inner xsep=0pt, tib] (t1s) {{\tib ་}};
\node[tgal=alB, right=0pt of t1s, tib] (t2) {{\tib ཀུན}};
\node[tgno, right=0pt of t2, inner xsep=0pt, tib] (t2s) {{\tib ་}};
\node[tgal=alA, right=0pt of t2s, tib] (t3) {{\tib དབང་དུ་མཛད་པ}};
\node[tgno, right=0pt of t3, inner xsep=0pt, tib] (t3s) {{\tib ་དེས། །}};
\node[tgno, right=4pt of t3s, tib] (t4) {{\tib ཁྱེད}};
\node[tgno, right=0pt of t4, inner xsep=0pt, tib] (t4s) {{\tib ་ལ་}};
\node[tgal=alE, right=0pt of t4s, tib] (t5) {{\tib བདེ་ལེགས}};
\node[tgno, right=0pt of t5, inner xsep=0pt, tib] (t5s) {{\tib ་}};
\node[tgal=alF, right=0pt of t5s, tib] (t6) {{\tib མཛད་པར་འགྱུར}};
\node[tgno, right=0pt of t6, inner xsep=0pt, tib] (t6s) {{\tib ། །}};
\draw[lk=alA] (b1.south) to[out=-90,in=90] (t3.north);
\draw[lk=alB] (b2.south) to[out=-90,in=90] (t2.north);
\draw[lk=alC] (b3.south) to[out=-90,in=90] (t1.north);
\draw[lk=alE] (b6.south) to[out=-90,in=90] (t5.north);
\draw[lk=alF] (b7.south) to[out=-90,in=90] (t6.north);
\end{scope}
\end{tikzpicture}
\end{center}
\caption{\label{fig:alignment} Two sentence pairs labeled by the teacher model (Section~\ref{sec:seeds}): the Sanskrit is segmented and lemmatized with ByT5-Sanskrit, and each content word is linked to the span of the translation that renders it, while function words such as \sa{ca} and pronouns such as \sa{sa} and \sa{va\d{h}} remain unlinked.}
\end{figure}

The lexica are built from existing components: the ByT5-Sanskrit analyzer \cite{nehrdich2024byt5}, which resolves sandhi, splits compounds and lemmatizes, the MITRA Qwen3.5-9B-Base stage 2 language model, adapted to these languages by continued pretraining \cite{nehrdich2026mitramt}, and a recipe of fine-tuning this model on 9{,}366 examples labeled by a stronger commercial teacher model.
What we contribute is the resource, a documented and reproducible way of building it, and a limited evaluation of how reliable it is:
\begin{itemize}
\item \textbf{MITRA-lexicon}: corpus-grounded lexica with span annotations for three language pairs with Sanskrit as source including Sanskrit--English, Sanskrit--Tibetan and Sanskrit--Chinese, each in both directions, with 2.18M headword entries, frequency-ranked senses, and 10.4M total attestations, every one with citation and highlight offsets. To our knowledge, this is the first resource of this kind for these language directions (Section~\ref{sec:resource}).
\item \textbf{The related data}: word alignments with quality signals for all sentence pairs, the aligner as an open-weights model, and its training data (Section~\ref{sec:method}).
\item \textbf{An assessment of reliability} against the reference dictionaries of the field for all three pairs and on expert-corrected gold alignments for Sanskrit--English, together with an inter-annotator agreement study in which three Sanskritists align the same 100 sentences independently. The gold alignments and the agreement data are released as well (Section~\ref{sec:eval}).
\item \textbf{A live web search interface} for looking up either language of a pair, with every sense linked to its attestations in the Dharmamitra corpus.
\end{itemize}
We release the lexica, alignments, gold data and evaluation protocol at \url{https://github.com/dharmamitra/dharmamitra-lexicon}, the interactive interface at \url{https://lexicon.dharmamitra.org}, and the word alignment model at \url{https://huggingface.co/buddhist-nlp/qwen35-mitra-lexicon-annotation}.

Section~\ref{sec:related} reviews related work, Sections~\ref{sec:method} and~\ref{sec:mining} describe how the alignments are produced and turned into entries, Section~\ref{sec:resource} presents the lexica and Section~\ref{sec:eval} gives the evaluation.

\section{Related Work}
\label{sec:related}

Böhtlingk and Roth's Sanskrit-to-German \textit{Sanskrit-Wörterbuch} \cite{pw}, published in seven volumes 1855--1875, is the most extensively documented Sanskrit dictionary. The widely used English-language dictionaries of Monier-Williams \cite{mw1872,mw1899} drew on it, and Apte \cite{apte1890} acknowledges using its quotations and references.
These works are not short of textual evidence, since Böhtlingk-Roth alone carries some 800{,}000 source references in its Cologne digitization and Apte, in the digitization we use, some 44{,}000. However, these citations are given as abbreviated references to editions that are mostly outdated and hard to access, without the translation of the passage, and without the frequency of a word or its distribution over the genres of the corpus.
For Sanskrit--Tibetan, \newcite{negi1993} quotes sentence pairs but draws on a comparatively small and concentrated set of sources and gives no frequencies.
For Sanskrit--Chinese, \newcite{wogihara1979} records the Chinese renderings of a word with the abbreviated titles of the texts in which they occur but without the passages, and \newcite{hirakawa1997} lists equivalents without naming any sources.

Among the digital lexical resources for Sanskrit, the Cologne Digital Sanskrit Dictionaries\footnote{\url{https://www.sanskrit-lexicon.uni-koeln.de}} have made a large array of classical dictionaries searchable in high quality, and the Sanskrit Heritage platform and its lexicon \cite{huet2005,platform12}\footnote{\url{https://sanskrit.inria.fr}} tie lexical entries to a generative morphology.
For Tibetan, Christian Steinert has created an interactive, searchable application for Tibetan lexical resources,\footnote{\url{https://dictionary.christian-steinert.de}} which also hosts the digital version of \newcite{negi1993} that we use in this study.
The Digital Corpus of Sanskrit (DCS\footnote{\url{https://www.sanskrit-linguistics.org/dcs/}}) \cite{hellwig-dcs} links a lemmatized corpus to dictionary entries and so offers attestations for Sanskrit words, though not their translations on sentence-level. The Sanskrit Sembank \cite{hellwig2025sembank} annotates occurrences in the DCS with word senses that are linked to the definitions of Monier-Williams.
\newcite{hellwig2026wsd} train a word sense disambiguation system on it and find that Monier-Williams's overlapping and periphrastic definitions are hard to discriminate, most of all for secondary senses. Among the ways of obtaining further sense annotations they single out alignment with translations, which has already served to source training data for word sense disambiguation in Ancient Greek \cite{keersmaekers2023wsd}.
MITRA-lexicon provides such alignments for Sanskrit at corpus scale, although it records translation equivalents rather than sense labels.
Large language models have also been used to generate lexicographic data directly, as for the Mangalam dictionary of Buddhist Sanskrit \cite{lugli2025mangalam}.

To keep our data compatible with the DCS, we process the Sanskrit side with ByT5-Sanskrit \cite{nehrdich2024byt5}, a byte-level sequence-to-sequence model trained jointly on segmentation, lemmatization and morphosyntactic tagging with the DCS as its supervision. It is robust to the orthographic variation of real-world e-texts and needs no external lexicon. Its output is, however, a silver annotation, and its errors propagate into the lexicon. Our headwords are therefore lemmas in the sense of the DCS and can be linked to the DCS and its corpus (Section~\ref{sec:refdict} examines how far this holds and how well the reference dictionaries cover the same lemmas).

Extracting translation lexica from parallel text is a classical application of word alignment \cite{gale1991identifying,melamed2000models}, and \newcite{melamed2000models} already evaluated such lexica against a published bilingual dictionary, as we do in Section~\ref{sec:refdict}.
Statistical aligners such as fast\_align \cite{dyer2013fastalign} and eflomal \cite{ostling2016eflomal} estimate lexical translation probabilities over token sequences, whereas SimAlign \cite{jalilisabet2020simalign} and awesome-align \cite{dou2021awesome} extract alignments from the similarity of contextual embeddings.
Supervised neural aligners fine-tune a multilingual encoder on gold alignments: as span prediction in the manner of extractive question answering \cite{nagata2020span}, with weakly supervised pretraining on Wikipedia entity links to reduce the need for manual labels \cite{wu2023wspalign}, or as binary sequence labeling \cite{latouche2024binaryalign}.
These supervised aligners presuppose gold alignments and a tokenized target, and a neural pretrained encoder that covers the languages.

Closest to the aligner we use is SemiAlign \cite{miao2025semialign}, a recent semi-supervised framework that, among other things, uses decoder-only LLMs as word aligners. Its authors fine-tune 8B-parameter LLMs with LoRA on a few hundred gold-aligned sentences and find that this alone does not reach the encoder-based state of the art, but that adding pseudo-labels produced by an encoder aligner for 30{,}000 to 52{,}000 unlabeled sentence pairs closes the gap and surpasses it on three of their four language pairs. We share their central observation that an LLM becomes a strong aligner once it is given enough automatically labeled data, but in our setting neither ingredient of their recipe exists. There are no gold alignments for Sanskrit, and no encoder aligner that could serve as pseudo-labeler for sandhied Sanskrit, unsegmented Tibetan or Buddhist Chinese. All of our supervision therefore comes from the combination of the ByT5-Sanskrit analyzer and a stronger teacher LLM prompted under a precision protocol.

Closest to our Sanskrit--English setting, Padamitra \cite{jagadeeshan2026padamitra} fine-tunes LLMs to generate word-level glossaries from verse--translation pairs, modeled on the word-by-word glosses of the \sa{R\=am\=aya\d{n}a} and the \sa{Bh\=agavata Pur\=a\d{n}a}.
The glosses of the Padamitra model are free text and are evaluated verse by verse, whereas we mark verbatim spans in the translation and aggregate them into attested, frequency-ranked dictionary entries.
The authors also identify the segmentation of sandhi and compounds as the main source of error, which corresponds to the dependence of our pipeline on the analyzer noted above.

Training a small, task-specific model on labels produced by a large commercial model is now a common paradigm \cite{hsieh2023distilling,zhou2024universalner}, see \newcite{tan2024annotation} for an overview. We use it for reasons of cost, as annotating enough word spans by hand to fine-tune a model would occupy highly trained annotators for a prohibitive amount of time.

\section{Methodology}
\label{sec:method}
This section describes the parallel corpora (Section~\ref{sec:data}) and how the aligner is obtained: the analysis of the Sanskrit (Section~\ref{sec:enrich}), alignment as span projection (Section~\ref{sec:task}), labeling by a teacher model (Section~\ref{sec:seeds}) and the training of a student on its labels (Section~\ref{sec:model}).

\subsection{Source Corpora}
\label{sec:data}

Table~\ref{tab:data} summarizes the three parallel corpora. For Sanskrit--English we use Mitrasa\d{m}graha \cite{nehrdich2026mitrasamgraha}, a sentence-aligned corpus of Sanskrit texts with published English translations.
It spans Vedic, epic, classical and medieval Sanskrit across six domains (Veda, epic, pur\=a\d{n}a, religious scripture, poetry, and scholastic treatises) and the work of more than forty translators. The alignments are silver-quality, produced with BertAlign and manually screened at document level.
For Sanskrit--Tibetan and Sanskrit--Chinese we use sentence pairs from the MITRA-parallel mining pipeline \cite{nehrdich2026mitramt}, which extracts parallel passages between Sanskrit, Tibetan and Buddhist Chinese from the digital canons by embedding-based retrieval and sentence alignment. We use MITRA-parallel~v3, in which the sentence alignment has been post-corrected by a dedicated model. From all three corpora we leave out the 18 texts whose Sanskrit is wholly or partly a modern reconstruction from the Tibetan or Chinese translation.
\begin{table}[t]
\begin{center}
\small
\begin{tabular}{llrr}
\toprule
\bf Pair & \bf Corpus & \bf Sentence pairs & \bf Sanskrit words \\
\midrule
Sanskrit--English & Mitrasa\d{m}graha & 395{,}528 & 5.87M \\
Sanskrit--Tibetan & MITRA-parallel v3 & 566{,}068 & 5.57M \\
Sanskrit--Chinese & MITRA-parallel v3 & 340{,}691 & 4.48M \\
\bottomrule
\end{tabular}
\end{center}
\caption{\label{tab:data} Parallel corpora used for word alignment in this work. Sanskrit words are counted after segmentation with ByT5-Sanskrit, compound members counted separately.}
\end{table}

\subsection{Sanskrit Analysis}
\label{sec:enrich}

In digital Sanskrit texts a sentence is usually not a sequence of words, as sandhi merges neighboring words, compounds fuse several lexical units into one orthographic word, and editions and digitizations follow different conventions when it comes to printing Sanskrit text. To standardize the input, we analyze the Sanskrit side of every sentence pair with ByT5-Sanskrit \cite{nehrdich2024byt5}, which yields for each word its unsandhied surface form, its lemma and a morphosyntactic tag. Non-final members of compounds are marked by a trailing hyphen (\sa{sarva-} \sa{ratn\=ani} in Figure~\ref{fig:alignment}).
Since the analyzer degrades on inputs beyond about 200 characters, we analyze long sentences in chunks of at most 150 characters and cut at white space where possible. The analyzer is not error-free, and a wrong lemma produces an entry under a wrong headword. The aggregation over the corpus pushes such errors into the low-frequency tail of the lexicon, but does not remove them.

\subsection{Alignment as Span Projection}
\label{sec:task}

We formulate word alignment as a tagging task over one sentence pair (Figure~\ref{fig:task}).
The analyzed Sanskrit words are wrapped in numbered tags, \tg{1}\ldots\tgc{1}, \tg{2}\ldots\tgc{2}, the same numbering is repeated over the lemmas, and the model must reproduce the target sentence verbatim while inserting each number around the target span that translates the corresponding word. Numbers without a confident equivalent are omitted. The model thus finds the spans in unsegmented target text itself, since no reliable segmenters for Tibetan and Chinese that match Sanskrit word boundaries are available. The verbatim constraint makes every output mechanically checkable, since the output with tags removed must equal the input and every tagged span must occur in it literally.
\begin{figure}[t]
\begin{center}
\fbox{\begin{minipage}{0.97\textwidth}\scriptsize
{\footnotesize\textbf{(a) Sanskrit--English}}\\[2pt]
\texttt{SANSKRIT: <1>rājā</1> <2>tu</2> <3>tam</3> <4>brāhmaṇam</4> <5>dṛṣṭvā</5> <6>prahṛṣṭaḥ</6> <7>abhavat</7>}\\
\texttt{LEMMAS:~~ <1>rājan</1> <2>tu</2> <3>tad</3> <4>brāhmaṇa</4> <5>dṛś</5> <6>prahṛṣ</6> <7>bhū</7>}\\
\texttt{ENGLISH:~ But the king, seeing that Brahmin, was filled with joy.}\\
\texttt{OUTPUT:~~ But the <1>king</1>, <5>seeing</5> that <4>Brahmin</4>, <6>was filled with joy</6>.}\\[6pt]
{\footnotesize\textbf{(b) Sanskrit--Chinese}}\\[2pt]
\texttt{SANSKRIT: <1>su</1> <2>jāti-</2> <3>kula-</3> <4>gotra-</4> <5>saṃpannaḥ</5> <6>ca</6> <7>bhaviṣyati</7>}\\
\texttt{LEMMAS:~~ <1>su</1> <2>jāti</2> <3>kula</3> <4>gotra</4> <5>sampad</5> <6>ca</6> <7>bhū</7>}\\
\texttt{CHINESE:~ }{\zh 生生世世生賢善家尊貴人中。}\\
\texttt{OUTPUT:~~ }{\zh 生生世世生賢善}\texttt{<3>}{\zh 家}\texttt{</3>}{\zh 尊貴人中。}
\end{minipage}}
\end{center}
\caption{\label{fig:task} Alignment as span projection. The target line, ENGLISH in (a), and CHINESE in (b), is copied verbatim and numbered tags are inserted around the span translating each Sanskrit word. In (a), \sa{tu}, \sa{tam} and \sa{abhavat} remain untagged: ``but'' and ``that'' are function words, and the copula is absorbed into the periphrasis tagged under \sa{prah\d{r}\d{s}\d{t}a\d{h}}. In (b) the Chinese paraphrases the Sanskrit, and only the one unmistakable correspondence, \sa{kula} : {\zh 家}, is marked.}
\end{figure}

\subsection{Teacher Labeling}
\label{sec:seeds}

Training data is produced by a commercial LLM (Gemini Flash with extended reasoning enabled)\footnote{\texttt{gemini-3.6-flash} with a reasoning budget of 16{,}384 tokens for Sanskrit--Tibetan and Sanskrit--Chinese (July 2026). \texttt{gemini-3.8-flash} at its highest reasoning level for Sanskrit--English (September 2026).} on a random sample of each corpus.
Since the labels are destined for lexicography, the prompts favor precision over completeness: the model is to tag a word only if its equivalent is unmistakable, never to tag function words and pronouns, to ignore free renderings, and to tag compound members individually only if their equivalents are separable in the translation.
A labeled pair is kept only if the output reproduces the target verbatim, which the teacher does in 96--99.8\% of the calls, and it tags 52--64\% of the Sanskrit words (Table~\ref{tab:seeds}).

\begin{table}[t]
\begin{center}
\small
\begin{tabular}{lrrrrr}
\toprule
\bf Direction & \bf Teacher calls & \bf Verbatim & \bf Words / sent. & \bf Tagged & \bf Seeds kept \\
\midrule
Sanskrit--English & 1{,}200 & 99.8\% & 13.8 & 52.2\% & 1{,}000 \\
Sanskrit--Tibetan & 3{,}542 & 99.6\% & 9.5 & 63.8\% & 2{,}000 \\
Sanskrit--Chinese & 2{,}518 & 96.0\% & 11.4 & 56.7\% & 2{,}000 \\
\midrule
Chinese--Tibetan (aux.) & 2{,}522 & 99.2\% & 11.4 & 44.0\% & 2{,}000 \\
English--Tibetan (aux.) & 2{,}469 & 99.5\% & 17.8 & 27.1\% & 1{,}786 \\
\bottomrule
\end{tabular}
\end{center}
\caption{\label{tab:seeds} Teacher labeling. ``Tagged'' is the mean fraction of source words that received a span. The two auxiliary directions are part of the training mixture of the student (Section~\ref{sec:model}) but their lexica are not the subject of this paper.}
\end{table}

\subsection{Student Model}
\label{sec:model}

The student is MITRA Qwen3.5-9B-Base stage 2 \cite{nehrdich2026mitramt}, the outcome of continued pretraining of Qwen3.5 \cite{qwen35} on Sanskrit, Tibetan, Buddhist Chinese and P\=ali. The student receives a one-line instruction and the data block of Figure~\ref{fig:task}, and is trained to emit the tagged target, which keeps inputs short.
All directions, including two auxiliary directions without Sanskrit (Table~\ref{tab:seeds}), are trained as one mixture (9{,}366 examples, with the Sanskrit--English seeds duplicated) by full-parameter fine-tuning on one 96\,GB GPU. Per direction, 50 seeds are held out for validation and 20 for testing.
To measure how faithfully the student reproduces the teacher, we labeled a further 1{,}489 sentence pairs (about 500 per direction, disjoint from the training seeds) with the teacher under the same protocol and scored the student's spans against them. A span counts as correct if it carries the same number and the same text.
The student reaches a span-level $F_1$ of 82.3 for Sanskrit--Tibetan, 87.4 for Sanskrit--Chinese and 86.2 for Sanskrit--English, each direction scored with the checkpoint that aligned its corpus.
This measures fidelity to the teacher, not correctness, and Section~\ref{sec:ablation} sets it against the agreement of the teacher with itself.

\section{Mining the Lexica}
\label{sec:mining}

\subsection{Corpus-Scale Inference}
\label{sec:inference}

The student is run over the three corpora with vLLM \cite{kwon2023vllm} using greedy decoding, and the mining of all three directions took about 36 GPU-hours on NVIDIA RTX PRO 6000 GPUs.
Every output is checked on signals that require no gold standard: whether it reproduces the target verbatim once the tags are removed, whether the tag numbering is well-formed, the mean log-probability of the generated tokens, and the share of tagged spans that do not occur in the target (ungrounded spans).
Table~\ref{tab:mining} shows that 96--99\% of the outputs are verbatim copies, and the fraction of ungrounded spans is below 0.3\% in every direction.
The student has inherited the teacher's restraint and leaves 30--50\% of the Sanskrit words unaligned, most for Sanskrit--Chinese and fewest for Sanskrit--Tibetan. 
\begin{table}[t]
\begin{center}
\small\setlength{\tabcolsep}{4.5pt}
\begin{tabular}{lrrrrr}
\toprule
\bf Direction & \bf Sent.\ pairs & \bf Verbatim & \bf Ungrounded & \bf Coverage & \bf Links \\
\midrule
Sanskrit--English & 395{,}528 & 97.4\% & 0.20\% & 0.52 & 2{,}952{,}451 \\
Sanskrit--Tibetan & 566{,}068 & 95.9\% & 0.01\% & 0.70 & 3{,}634{,}841 \\
Sanskrit--Chinese & 340{,}691 & 98.8\% & 0.02\% & 0.50 & 1{,}942{,}754 \\
\midrule
Total & 1{,}302{,}287 & & & & 8{,}530{,}046 \\
\bottomrule
\end{tabular}
\end{center}
\caption{\label{tab:mining} Corpus-scale inference. Verbatim: share of outputs that reproduce the target exactly once the tags are removed. Ungrounded: mean share of the tagged spans of an output that do not occur in the target. Coverage: mean fraction of Sanskrit words per sentence that received a grounded span.}
\end{table} For Sanskrit--English, outputs that are not verbatim, have malformed numbering, a mean log-probability below $-0.45$ or more than 25\% ungrounded spans are excluded from the lexicon, since in each of these cases the output itself is unreliable. This removes 17{,}068 rows (4.3\%). For Sanskrit--Tibetan and Sanskrit--Chinese, whose sentence pairs have been post-corrected and verified by a dedicated model in MITRA-parallel~v3, all rows are used that render the target verbatim. 

\subsection{From Alignments to Lexicon Entries}
\label{sec:build}

For every aligned Sanskrit word we record the character offsets of its equivalent in the target and, by fuzzy string matching between the unsandhied forms and the source string, of the word in the original sentence.
Word spans are keyed on the lemma and constitute the single-word lexicon. Phrase spans are built by consecutive members of a compound that are all aligned and where their equivalents form one unit in the target, separated at most by case particles in Tibetan, or by glue words such as \textit{of} and \textit{the} in English, where the members may also occur in any order, the compound is emitted under its hyphenated members, with the final member reduced to its stem as a lemma would be (\sa{sarva-ratna} for \sa{sarva-ratn\=ani}), together with its sandhied surface form, so that it can be looked up exactly as it stands in the text. Only the full compound is emitted, not its parts, whose members remain available as word entries.
In Figure~\ref{fig:alignment}a, for instance, the links yield the word entries \sa{h\d{r}} : ``carried away'', \sa{tejas} : ``force'' and \sa{ratna} : ``jewels'', and the compound entry \sa{sarva-ratna} : ``all the jewels''.
Table~\ref{tab:spans} gives the quantitative overview.

For Chinese, where the compound members are usually rendered by adjacent characters, a rule that admits no intervening material and no reordering misses many compounds that are rendered with the members in another order (\sa{pa\~nca-bhaya-samatikr\=anta} : {\zh 離五怖} `free from the five fears'), with a grammatical particle between them (\sa{n\=aga-r\=aja-parip\d{r}cch\=a} : {\zh 龍王所問}) or with one member left untranslated (\sa{cint\=a-maya-praj\~n\=a} : {\zh 思慧}).
We therefore admit for Chinese any order of the members within one contiguous stretch, a single intervening character if it is one of fourteen particles ({\zh 之所而為其及與於諸能得等者中}), and one unaligned member inside the compound, but no punctuation and no other intervening material. This adds 64{,}757 phrase spans (Table~\ref{tab:spans}).
Two further rules form compounds whose members the labeling protocol leaves incomplete. Abstract and delimiting suffixes (\sa{-tva}, \sa{-t\=a}, \sa{-m\=atra}) are analyzed as members of their own and usually remain untagged, so that \sa{dharma-t\=a} would yield no compound entry. When the other members of such a compound are aligned as one unit, we emit the full compound and extend its span over a directly following \textit{nyid} or \textit{tsam} in Tibetan and {\zh 性} in Chinese (\sa{dharma-t\=a} : \textit{chos nyid}, {\zh 法性}). For Chinese, a compound of three or more members whose equivalents are separated only by the enumeration comma {\zh 、} is emitted as one unit (\sa{r\=aga-dve\d{s}a-moha-prah\=a\d{n}a} : {\zh 斷除貪、瞋、癡}); with two members this rule proved unreliable and is not applied. In samples checked by hand, 28 of 30 Tibetan and 23 of 25 Chinese suffix compounds and all 25 enumerations were correct. The two rules add 10{,}109 compounds for English, 29{,}737 for Tibetan and 17{,}172 for Chinese.

\begin{table}[t]
\begin{center}
\small
\begin{tabular}{lrrr}
\toprule
\bf Direction & \bf Word spans & \bf Phrase spans & \bf Total \\
\midrule
Sanskrit--English & 2{,}749{,}903 & 566{,}022 & 3{,}315{,}925 \\
Sanskrit--Tibetan & 3{,}634{,}045 & 1{,}090{,}654 & 4{,}724{,}699 \\
Sanskrit--Chinese & 1{,}942{,}753 & 532{,}475 & 2{,}475{,}228 \\
\bottomrule
\end{tabular}
\end{center}
\caption{\label{tab:spans} Spans extracted from the alignments. Phrase spans comprise full compounds and their sandhied surface forms.}
\end{table}


Three filters remove noise: 1) We filter senses whose gloss crosses a clause boundary marked by punctuation, or whose pointwise mutual information \cite{church1990pmi} with the headword is below 1.0, removing glosses that contribute little to the relative attestation of a headword. 2) We remove Tibetan glosses that consist only of function syllables and attach indiscriminately to many headwords. 3) Function words and pronouns are filtered, since the labeling protocol already excludes them (Section~\ref{sec:seeds}) and the student follows it.
Headwords that reproduce typing errors of the e-texts are removed where these can be recognized: a headword is dropped when one of its members occurs no more than five times as a lemma or form in the Sanskrit collection of DharmaNexus (some 73 million words, analyzed with the same analyzer), lies one edit away from a lemma or form that occurs at least 300 times there, and is not a headword of Monier-Williams or Apte (\textit{kre\'sa} for \sa{kle\'sa}, \textit{\'smatha} for \sa{\'samatha}, \textit{pu\d{n}dar\={\i}ka} for \sa{pu\d{n}\d{d}ar\={\i}ka}).
This removes 35{,}075 Sanskrit headwords (3.0\%) with 51{,}420 attestations (0.5\%), 91\% of them attested once or twice.

\section{The MITRA-lexicon Resource}
\label{sec:resource}

MITRA-lexicon consists of three lexicon pairs with Sanskrit as the source language, each released in both directions.
\label{sec:entry}
\begin{figure}[t]
\begin{center}
\fbox{\begin{minipage}{0.97\textwidth}\scriptsize
\texttt{\{"sa": "kṛpā", "n": 604, "n\_senses": 38,}\\
\hspace*{0.6em}\texttt{"senses": [}\\
\hspace*{1.1em}\texttt{\{"en": "mercy", "count": 151,}\\
\hspace*{1.7em}\texttt{"ex": [\{"id": "train/cc-by-sa/GE07hv\_conpu.tsv:1716", "license": "cc-by-sa",}\\
\hspace*{6.1em}\texttt{"src": "kṛpayā tac ca jagrāha śaṃtanur mṛgayāṃ gataḥ /",}\\
\hspace*{6.1em}\texttt{"target": "Going out a-hunting the king Shāntanu, out of mercy, took them up.",}\\
\hspace*{6.1em}\texttt{"start": 46, "end": 51\}, ...]\},}\\
\hspace*{1.1em}\texttt{\{"en": "compassion", "count": 110, "ex": [...]\},}\\
\hspace*{1.1em}\texttt{\{"en": "pity", "count": 86, "ex": [...]\}, ...]\}}
\end{minipage}}
\end{center}
\caption{\label{fig:entry} An entry of the Sanskrit$\rightarrow$English lexicon (abridged). \texttt{n} is the number of attestations of the headword, \texttt{count} that of a sense; \texttt{start} and \texttt{end} are character offsets of the equivalent in \texttt{target}. The identifier encodes the portion of the corpus, the license group, the text and the line.}
\end{figure}
An entry (Figure~\ref{fig:entry}) consists of a headword, its total number of attestations, and a list of senses.
A sense is a translation equivalent as it occurs in the corpus, with the number of times the headword was rendered in this way, and senses are ranked by this count.
Each sense carries all of its attestations, one for each of these occurrences: the identifier of the sentence in its source corpus, the text of the translation, and the character offsets of the equivalent within it, so that an interface can highlight it. Headwords are of the three kinds produced in Section~\ref{sec:build}.
Lemma entries use DCS lemmas as keys, combining the evidence for \sa{k\d{r}p\=a}, \sa{k\d{r}pay\=a}, \sa{k\d{r}p\=am} and their sandhi variants in one entry. Compound entries use the hyphenated, unsandhied members of a compound as keys, with the final member reduced to its stem, and the case forms that occur are listed with the entry (in the Sanskrit--Chinese lexicon, \sa{ku\'sala-m\=ula} combines 1{,}514 attestations of 16 forms).
Surface entries index the same compounds under the sandhied string as it stands in the text, located by fuzzy string matching.
In the reverse lexica (English, Tibetan or Chinese $\rightarrow$ Sanskrit) the headwords are the equivalents and the senses are Sanskrit lemmas and compounds.

\subsection{Size and Coverage}
\label{sec:size}

\begin{table}[t]
\begin{center}
\small
\begin{tabular}{llrrrr}
\toprule
\bf Pair & \bf Direction & \bf Headwords & \bf Senses & \bf Senses / headword & \bf Attestations \\
\midrule
Sanskrit--English & sa$\rightarrow$en & 418{,}927 & 950{,}845 & 2.27 & 3{,}293{,}993 \\
                  & en$\rightarrow$sa & 396{,}289 & & 2.40 & \\
Sanskrit--Tibetan & sa$\rightarrow$bo & 529{,}599 & 895{,}229 & 1.69 & 4{,}665{,}998 \\
                  & bo$\rightarrow$sa & 445{,}176 & & 2.01 & \\
Sanskrit--Chinese & sa$\rightarrow$zh & 223{,}081 & 422{,}074 & 1.89 & 2{,}447{,}898 \\
                  & zh$\rightarrow$sa & 171{,}780 & & 2.46 & \\
\midrule
Total & & 2{,}184{,}852 & 2{,}268{,}148 & & 10{,}407{,}889 \\
\bottomrule
\end{tabular}
\end{center}
\caption{\label{tab:resource} The released lexica. Senses (headword--gloss pairs) and attestations are properties of a pair and identical in both of its directions. Headwords comprise lemmas as well as compound and phrase entries.}
\end{table}

Table~\ref{tab:resource} describes the six directional lexica.
In each pair there are fewer headwords on the target side than on the Sanskrit side, and correspondingly more senses per target headword.
For Tibetan (2.01 against 1.69) and Chinese (2.46 against 1.89) this reflects translation vocabularies that are smaller and more standardized than the Sanskrit they render. That English shows the same pattern (2.40 against 2.27) indicates that many near-synonyms of the Sanskrit lexicon converge on the same English word in translation practice.

\begin{table}[t]
\begin{center}
\small
\begin{tabular}{lrrrrrrr}
\toprule
\bf Attestations & \bf 1 & \bf 2--4 & \bf 5--19 & \bf 20--99 & \bf 100--999 & \bf $\geq$1000 & \bf Total \\
\midrule
Sanskrit--English & 159{,}756 & 29{,}615 & 12{,}918 & 6{,}511 & 3{,}382 & 491 & 212{,}673 \\
Sanskrit--Tibetan & 124{,}176 & 60{,}734 & 17{,}776 & 6{,}397 & 3{,}159 & 614 & 212{,}856 \\
Sanskrit--Chinese & 42{,}567 & 29{,}254 & 10{,}120 & 4{,}041 & 2{,}008 & 350 & 88{,}340 \\
\bottomrule
\end{tabular}
\end{center}
\caption{\label{tab:freq} Single-word Sanskrit headwords by number of attestations. The remaining headwords of Table~\ref{tab:resource} are compound and surface entries.}
\end{table}

In accordance with Zipf's law \cite{zipf1949}, the distribution of our headwords is heavily skewed (Table~\ref{tab:freq}).
Of the 212{,}673 single-word Sanskrit headwords of the English lexicon, 75\% are attested once.
The core of each lexicon, which we define as single-word headwords with at least five attestations, comprises 23{,}302 headwords for English, 27{,}946 for Tibetan and 16{,}519 for Chinese.
In each pair this core accounts for more than 90\% of the attestations of single-word headwords.
To assess rare headwords, we reviewed 20 randomly drawn entries with a single attestation for each pair of the released lexica. Fifty-five of the 60 are compounds, phrases or sandhied surface forms, nearly all of them compositional rather than lexicalized.
None was misaligned or drawn from a mismatched sentence pair. Thirty-nine were correct translation equivalents (16 for English, 10 for Tibetan, 13 for Chinese), such as \sa{vy\=aghra-carma-dhara} : ``wore a tiger's skin'', \sa{m\=a\d{n}\d{d}aleya-devat\=a-pari\d{s}ad} : \textit{dkyil 'khor gyi lha'i 'khor} or \sa{phala-upabhojaka} : {\zh 受用果}, and 15 were correct but imprecise (3, 7, 5): in nine of them a surface key also takes in a neighboring word that is not part of the alignment (a shortcoming of the fuzzy matching), in three the span is a word too wide or too narrow, and in three the translator leaves a member of the compound unrendered.
The 6 other errors (1, 3, 2) all lie on the Sanskrit side, in the key or the lemma, while the aligned translation is correct in all of them: a particle or a finite verb fused into the key (\textit{hakar\d{s}a\d{n}a} for \sa{ha kar\d{s}a\d{n}\=at}, \textit{sarva-buddha-dadanti} for \sa{sarva-buddh\=a dadanti}), a word cut at a sandhi boundary, a typing error of the e-text carried into the key (\textit{tathat\=a-rupa} for \sa{tathat\=a-r\=upa}), mantra syllables analyzed as a word, and a punctuation mark carried into the key.

\subsection{Composition by Genre and Period}
\label{sec:genre}

\begin{table}[t]
\begin{center}
\small
\begin{tabular}{lrrrrrr}
\toprule
\bf Genre & \bf Vedic & \bf Epic & \bf Classical & \bf Medieval & \bf Total & \bf \% \\
\midrule
Veda                & 339{,}419 &           &             &           & 339{,}419   & 10.3 \\
Epic                &           & 634{,}817 &             &           & 634{,}817   & 19.3 \\
Pur\=a\d{n}a        &           &           & 1{,}187{,}279 &         & 1{,}187{,}279 & 36.0 \\
Religious scripture &           & 58{,}196  & 231{,}568   & 61{,}703  & 351{,}467   & 10.7 \\
Treatises           &           & 34{,}261  & 530{,}746   & 14{,}993  & 580{,}000   & 17.6 \\
Poetry              &           &           & 26{,}013    & 174{,}998 & 201{,}011   & 6.1 \\
\midrule
Total               & 339{,}419 & 727{,}274 & 1{,}975{,}606 & 251{,}694 & 3{,}293{,}993 & 100 \\
\bottomrule
\end{tabular}
\end{center}
\caption{\label{tab:genre} Attestations of the Sanskrit--English lexicon, by genre and period of the Sanskrit text as reported in \protect\newcite{nehrdich2026mitrasamgraha}. Periods are Vedic before 500 BCE, Epic 500 BCE--200 CE, Classical 200--1000 CE, Medieval 1000--1800 CE.}
\end{table}

For Sanskrit-English, by genre (Table~\ref{tab:genre}), pur\=a\d{n}as contribute over a third of the attestations, scholastic treatises and epic about a fifth each, religious scripture and the Veda 10\% each and poetry 6\%. By period, 60\% of the attestations are classical, 22\% epic, 10\% Vedic and 8\% medieval.
Two texts, the seven-kha\d{n}\d{d}a (vulgate) \sa{Skandapur\=a\d{n}a} in Tagare's translation (830{,}897 attestations) and the \sa{Mah\=abh\=arata} (576{,}228), together account for 43\% of the attestations of the Sanskrit--English lexicon, and their translators' conventions are correspondingly prominent among the top-ranked English senses of common words.
Table~\ref{tab:collection} breaks down the Tibetan and Chinese attestations by the DharmaNexus category of the Sanskrit text. For Tibetan, treatises provide 53\% of the attestations, led by tantric commentaries, Yog\=ac\=ara, Abhidharma, Madhyamaka and Pram\=a\d{n}a. For Chinese, scriptures provide 72\%, the s\=utras and the Praj\~n\=ap\=aramit\=a about a fifth each, and among the treatises Yog\=ac\=ara and Abhidharma dominate, while the Pram\=a\d{n}a and tantric treatises that are prominent for Tibetan are nearly absent.

\begin{table}[t]
\begin{center}
\small
\begin{tabular}{llrrrr}
\toprule
 & & \multicolumn{2}{c}{\bf Sanskrit--Tibetan} & \multicolumn{2}{c}{\bf Sanskrit--Chinese} \\
\bf Collection & \bf Category & \bf Attestations & \bf \% & \bf Attestations & \bf \% \\
\midrule
Scriptures & S\=utra                        & 602{,}459 & 12.9 & 541{,}232 & 22.1 \\
           & Praj\~n\=ap\=aramit\=a         & 708{,}552 & 15.2 &  536{,}847 &  21.9 \\
           & Tantra                         & 431{,}906 &  9.3 & 208{,}568 &  8.5 \\
           & Vinaya                         & 338{,}486 &  7.3 &  163{,}853 &  6.7 \\
           & Other                          &  126{,}426 &  2.7 & 308{,}665 & 12.6 \\
Treatises  & Tantra                         & 647{,}564 & 13.9 &       6{,}809 &  0.3 \\
           & Yog\=ac\=ara                   & 496{,}394 & 10.6 & 355{,}544 & 14.5 \\
           & Madhyamaka                     & 388{,}868 &  8.3 &  86{,}589 &  3.5 \\
           & Abhidharma                     & 407{,}239 &  8.7 & 210{,}367 & 8.6 \\
           & Pram\=a\d{n}a                  & 239{,}638 &  5.1 &   1{,}392 &  0.1 \\
           & Other                          & 278{,}466 &  6.0 &  28{,}032 &  1.1 \\
\midrule
Total      &                                & 4{,}665{,}998 & 100 & 2{,}447{,}898 & 100 \\
\bottomrule
\end{tabular}
\end{center}
\caption{\label{tab:collection} Attestations of the Sanskrit--Tibetan and Sanskrit--Chinese lexica by the DharmaNexus collection and category of the Sanskrit text. Other scriptures are Ratnak\=u\d{t}a, Avata\d{m}saka (for Chinese 219{,}201 attestations) and miscellaneous s\=utras. Other treatises include hymns, Praj\~n\=ap\=aramit\=a and s\=utra commentaries, letters, J\=ataka, Vinaya commentary and n\={\i}ti.}
\end{table}

\subsection{Access}

We release the six lexica as line-delimited JSON, the span files from which they were built, the alignment output for every sentence pair with its quality signals, the teacher-labeled seeds, and the student model.
Since every attestation reproduces its sentence pair, the Sanskrit--English lexicon inherits the licenses of the texts of Mitrasa\d{m}graha and is, like the corpus, released in two versions: one under CC BY-SA 4.0 that contains the attestations from texts under CC BY-SA (92.3\% of all attestations), and the complete lexicon under CC BY-NC-ND 4.0, which adds those from texts under that license (4.3\%) and from the validation and test splits of Mitrasa\d{m}graha, for which no license is stated (3.4\%), and is the version described in this paper. The Sanskrit--Tibetan and Sanskrit--Chinese lexica are released under CC BY-SA 4.0.

The lexica are served through a web interface backed by a search index with one document per term and one per attestation.
Lookup works from either language of a pair and is script-agnostic: Sanskrit may be typed with or without diacritics (\texttt{karuna} finds \sa{karu\d{n}\=a}), Tibetan in Tibetan script or in Wylie.
Suggestions are ranked by attestation frequency.
An entry lists its senses with counts, grouped by collection and text, and each citation displays the Sanskrit sentence and its translation with the word and its equivalent highlighted on both sides.

\section{Evaluation}
\label{sec:eval}
We evaluate the generated entries along two vectors: whether they are correct, and whether the word alignments from which they are aggregated are correct.
The first we answer for all three pairs with reference dictionaries (Section~\ref{sec:refdict}).
The second requires gold alignments by experts, which we could obtain only for Sanskrit--English, where several Sanskritists could be recruited (Section~\ref{sec:gold}).
Section~\ref{sec:baselines} compares the aligner with statistical aligners and with the teacher.

\subsection{Agreement with Reference Dictionaries}
\label{sec:refdict}

For every headword that a reference dictionary can judge, we ask whether the mined senses are equivalents that the reference records. As references for Sanskrit--English, we use the dictionary of \newcite{mw1872}, alone and in union with \newcite{apte1890}, both in the digitization of the Cologne Digital Sanskrit Dictionaries.
We do not use Böhtlingk and Roth's \textit{Sanskrit-Wörterbuch} \cite{pw}, the most comprehensive of the three, because its definitions are in German.
For Sanskrit--Tibetan, we use a digitized version of \newcite{negi1993}\footnote{From Christian Steinert's Tibetan dictionary collection at \url{https://dictionary.christian-steinert.de}, with the data at \url{https://github.com/christiansteinert/tibetan-dictionary} (accessed August 2026).}, which we invert from its Tibetan headwords into a Sanskrit$\rightarrow$Tibetan index.
For Sanskrit--Chinese, we use a digitized version of \newcite{hirakawa1997} that we produced ourselves.
Reference citation forms are mapped to the stems used as mined headwords by an explicit list of rules (\sa{dharma\d{h}} $\rightarrow$ \sa{dharma}, \sa{r\=aj\=a} $\rightarrow$ \sa{r\=ajan}, and so on).

We evaluate single-word Sanskrit headwords and let $H$ be the set of mined headwords that also occur as a Sanskrit item in the reference.
For $h\in H$ we report P@$k$, the proportion of headwords for which one of the $k$ top-ranked mined senses matches a reference equivalent; the attestation-weighted precision, the share of all attestations of the headwords in $H$ that fall on a matching sense; and, where the reference lists equivalents, the mean proportion of the reference equivalents of $h$ found among its mined senses (recall).
Matching is strict when the normalized strings are equal.
For Tibetan, normalization strips trailing case particles and clitics, and for Chinese it unifies character variants.
Because translators add and drop modifiers freely (\textit{snying rje chen po} for \textit{snying rje}), we also report a containing variant, in which the mined sense may contain a reference equivalent, and a relaxed variant with containment in either direction.
The last is lenient, since a short mined sense can be contained in a longer reference term by accident, and is given for orientation only.
For English the references are definitions in prose rather than lists of equivalents.
A mined gloss counts as matched (content-word matching) when all its content words, after stemming, occur in the text of the entry of exactly that headword, so recall is undefined.
Glosses that merely repeat the Sanskrit word (untranslated \sa{dharma}, \sa{karu\d{n}\=a}) can be excluded (``no loans'').

\begin{table}[t]
\begin{center}
\small
\begin{tabular}{llrrrrr}
\toprule
\bf Pair & \bf Reference, matching & $|H|$ & \bf P@1 & \bf P@5 & \bf Att.-weighted prec. & \bf Recall \\
\midrule
Sanskrit--Tibetan & Negi, strict & 24{,}858 & 58.2 & 71.4 & 76.6 & 46.0 \\
                  & Negi, containing & & 60.5 & 73.1 & 79.9 & 50.9 \\
                  & Negi, relaxed & & 68.0 & 78.6 & 85.7 & 60.9 \\
\midrule
Sanskrit--Chinese & Hirakawa, strict & 15{,}783 & 53.0 & 69.4 & 75.2 & 44.1 \\
                  & Hirakawa, containing & & 57.2 & 72.1 & 80.4 & 47.8 \\
                  & Hirakawa, relaxed & & 68.4 & 80.3 & 88.3 & 63.4 \\
\midrule
Sanskrit--English & MW & 33{,}869 & 34.2 & 49.9 & 63.0 & -- \\
                  & MW $\cup$ Apte & 35{,}107 & 36.6 & 52.0 & 67.3 & -- \\
                  & MW $\cup$ Apte, no loans & 32{,}304 & 40.9 & 56.0 & 69.0 & -- \\
\bottomrule
\end{tabular}
\end{center}
\caption{\label{tab:refdict} Agreement of the mined Sanskrit-source lexica with reference dictionaries (in \%), over all single-word headwords that the reference can judge. Strict and containing figures are lower bounds on precision (see text).}
\end{table}

\begin{table}[t]
\begin{center}
\small
\begin{tabular}{lrrrrrr}
\toprule
\bf Attestations of headword & \bf 1 & \bf 2--4 & \bf 5--19 & \bf 20--99 & \bf 100--999 & \bf $\geq$1000 \\
\midrule
Sanskrit--Tibetan, P@1 & 45.4 & 50.6 & 56.9 & 71.9 & 84.3 & 90.0 \\
Sanskrit--Tibetan, P@5 & 45.4 & 64.2 & 80.0 & 91.7 & 96.6 & 96.6 \\
Sanskrit--Chinese, P@1 & 39.9 & 42.3 & 49.3 & 65.7 & 77.0 & 88.1 \\
Sanskrit--Chinese, P@5 & 39.9 & 56.2 & 74.0 & 87.5 & 93.6 & 98.8 \\
Sanskrit--English, P@1 & 20.4 & 24.8 & 38.1 & 54.4 & 75.0 & 90.2 \\
Sanskrit--English, P@5 & 20.4 & 39.0 & 64.0 & 81.6 & 93.9 & 98.5 \\
\midrule
Headwords in $H$ (Tibetan) & 6{,}071 & 6{,}154 & 5{,}796 & 3{,}975 & 2{,}362 & 500 \\
Headwords in $H$ (Chinese) & 2{,}507 & 3{,}957 & 4{,}262 & 2{,}977 & 1{,}751 & 329 \\
Headwords in $H$ (English) & 9{,}532 & 8{,}511 & 7{,}965 & 5{,}488 & 3{,}130 & 481 \\
\bottomrule
\end{tabular}
\end{center}
\caption{\label{tab:bands} Agreement by frequency of the headword (strict matching for Tibetan and Chinese; MW $\cup$ Apte for English). With a single attestation there is a single sense, so that P@1 and P@5 coincide.}
\end{table}

The set $H$ comprises 12\% (Tibetan), 18\% (Chinese) and 17\% (English) of the mined single-word headwords, but 73\%, 85\% and 87\% of their attestations: the headwords that a reference cannot judge are overwhelmingly rare forms.
Table~\ref{tab:refdict} gives the results over all of $H$ and Table~\ref{tab:bands} by frequency of the headword.
Under strict matching, 77\% of the Sanskrit--Tibetan attestations, 75\% of the Sanskrit--Chinese and, under content-word matching, 63--69\% of the Sanskrit--English attestations fall on senses that the reference confirms.
Agreement rises steeply with frequency.
For headwords with at least 100 attestations, the top-ranked sense is confirmed in 84--90\% of the cases for Tibetan, 77--88\% for Chinese and 75--90\% for English, and one of the five top-ranked senses in 97\%, 94--99\% and 94--99\%.
For headwords attested once, agreement drops to 45\%, 40\% and 20\%.
As a control, we re-ran the evaluation with the reference entries permuted among headwords of the same frequency band: P@1 then falls below 1\% for all three pairs, so that chance agreement is negligible, including under the content-word matching used for English.

These figures are lower bounds on precision, and a reading of samples from both sides shows why.
Many mined senses that a reference lacks are plain renderings it does not record (\textit{bu mo} for \sa{duhit\d{r}}, where our index of Negi has only the honorific \textit{sras mo}, \textit{rin po che} for \sa{ratna}, {\zh 汝} for \sa{tvad}) or, for English, synonyms that a prose definition happens not to contain (home for \sa{g\d{r}ha}, arrogance against a definition reading ``pride, haughtiness''), a consequence of the clustered, near-synonymous definitions of Monier-Williams \cite{hellwig2026wsd}.
Target-side variation that our normalization does not unify counts as disagreement (case-marked forms such as \textit{mthus} beside \textit{mthu} for \sa{anubh\=ava}, abbreviated compounds such as \textit{stong nyid} for \textit{stong pa nyid}), and the digitized references are imperfect: our Hirakawa is an OCR transcription of the scanned print, and the 1872 edition of Monier-Williams nests derivatives and compounds in the entry of their base, so that sub-entries had to be split off heuristically.
Errors of alignment and analysis occur as well ({\zh 百} for \sa{var\d{s}a}, from {\zh 百歲} `a hundred years').
Negi and Hirakawa excerpt the same canonical translations from which our corpus is drawn, which is reflected in the higher agreement for Tibetan and Chinese, whereas the English references were not compiled from the translations of our corpus.
Conversely, the resource does not recover everything the lexicographers collected (recall of 46--61\% against Negi and 44--63\% against Hirakawa).
The references record the learned vocabulary of lexicographical and poetic literature (\textit{kha ba'i dgra bo} `enemy of snow' for \sa{agni}), further meanings of polysemous words (\textit{chang} for \sa{ma\d{n}\d{d}a}), rare transcriptions ({\zh 多他阿伽度} beside the mined {\zh 如來} for \sa{tath\=agata}) and explanatory renderings ({\zh 五百弓} `five hundred bow-lengths' for \sa{kro\'sa}), and some of their sources are not in our corpus.
Of 20 missing Negi equivalents of frequent headwords that we examined, seven are documented only in lexical works, six are artefacts of our parse of Negi or of the matching, five come from passages that our corpus lacks or has only in another reading, two occur in our corpus but were left unaligned by the student, and none is a misalignment.

Negi's citations also do not separate or mark evidence that has been reconstructed based on Tibetan or other sources, as the example of the  \textit{Madhy\=antavibh\=aga\d{t}\={\i}k\=a} shows. It is cited from the edition of \newcite{pandeya1971}, and 22 of its 224 citations quote passages that the previous editors restored from the Tibetan translation.
As our headwords follow the DCS, participles and absolutives are lemmatized to the verbal root, so evidence a lexicographer files under \sa{k\d{r}ta} is credited to \sa{k\d{r}} for example.
Both sides shrink with frequency: the share of reference equivalents without a mined counterpart falls from 57\% (Negi) and 59\% (Hirakawa) for headwords attested once to 18--25\% for headwords with at least 100 attestations, and the share of mined senses without a counterpart in the reference from 58--77\% of the senses attested once to less than 9\% of those attested at least 1{,}000 times.

The references also rest on a much narrower textual basis than our corpus.
Negi quotes about 73{,}000 distinct Sanskrit passages, against 565{,}642 sentence pairs in our Sanskrit--Tibetan lexicon, half of its text citations come from ten sources, and a seventh of his pairs are documented only in lexical works such as the Mah\=avyutpatti.
Half of Hirakawa's vocabulary, located in the Chinese canon, is concentrated in 76 texts, and three quarters of Apte's source references point to classical poetry and drama, whereas the epics and pur\=a\d{n}as make up more than half of our Sanskrit--English attestations.
Agreement with the references can therefore only be partial.

Many of Hirakawa's entries are phrases tied to a passage (28\% of his pairs have a Chinese of four or more characters), so we tested each of his 123{,}248 pairs with a Chinese of at least two characters in three ways (Table~\ref{tab:hirakawagap}): whether it is a sense of our lexicon, whether its Chinese and its Sanskrit co-occur in one of our sentence pairs, and whether it can be composed from our entries ({\zh 加行薩埵} from \sa{prayoga} : {\zh 加行} and \sa{sattva} : {\zh 薩埵}), with controls on neighboring sentence pairs and on other headwords succeeding for only 6\% and 1\%.
By these criteria 70\% of Hirakawa is contained in the resource or documented by its corpus.
Of the rest, 14 points are Chinese that occurs in none of our sentence pairs, often headwords written discontinuously or with notes, or OCR recognition errors, and 14 points are pairs whose Chinese and Sanskrit both occur in our corpus but never together.
Of 17 such pairs that we traced in detail, five are in our lexicon under a form the comparison does not recognize (\sa{paryavasthita} : {\zh 現起}, ours under \sa{paryavasth\=a}). Nine occur together in only one or two of our sentence pairs, where the aligner left the word unaligned or gave it a different or shorter span ({\zh 所生} for \sa{janya}, where Hirakawa has {\zh 所生法}), and three are artefacts of Hirakawa or lie in passages without a Sanskrit counterpart in our corpus.
\begin{table}[t]
\begin{center}
\small
\begin{tabular}{lr}
\toprule
Sense of the lexicon & 31.8 \\
Co-occurring in a sentence pair & 28.0 \\
Composable from entries & 9.8 \\
\quad \textit{covered} & \textit{69.7} \\
\midrule
Chinese not in our sentence pairs & 7.5 \\
Chinese not in the Chinese collection & 7.0 \\
Both in the corpus, never in one pair & 14.1 \\
Sanskrit not in our corpus & 1.8 \\
\bottomrule
\end{tabular}
\end{center}
\caption{\label{tab:hirakawagap} The 123{,}248 pairs of Hirakawa (Chinese of at least two characters) by the evidence of MITRA-lexicon (in \%, each pair in the first row that applies).}
\end{table}

The comparison is confined to headwords that a reference contains, and Table~\ref{tab:lemmacov} shows how many that are.
Our own headwords follow the DCS closely: of the single-word headwords with at least 100 attestations, 99\% (English), 98\% (Tibetan) and 97\% (Chinese) are DCS lemmas, the rest being mostly sandhied surface forms of compounds and, for Chinese, inflected forms that the analyzer left unlemmatized in the formulaic prose of the Praj\~n\=ap\=aramit\=a and the Vinaya (\textit{d\=anap\=aramit\=ay\=a\d{m}}).
Whether a lemma is found in a dictionary depends in part on citation conventions. The analyzer follows the DCS (\sa{bhagavant} where the dictionaries have \sa{bhagavat}, causative stems such as \sa{de\'say}, roots with preverbs such as \sa{upalabh}), and Negi's Sanskrit evidence hardly covers any verbal roots.
For Sanskrit--English, Monier-Williams and Apte contain our frequent headwords about as often as DCS lemmas of comparable frequency (89\% against 86\%).
The Buddhist pairs show a gap of their own, and about a quarter of the frequent Sanskrit headwords of our Sanskrit--Tibetan and Sanskrit--Chinese lexica are found neither in Monier-Williams nor in Apte.
Edgerton's dictionary of Buddhist Hybrid Sanskrit \cite{edgerton1953}, which by design records only what departs from the classical language, covers about a quarter and a third of these gaps (632 of 2{,}270 and 439 of 1{,}367 headwords), and 17\% and 15\% of the frequent headwords remain without an entry in any of the three. This shows an important gap for future work on the DCS and for this lexical resource, where the annotation of non-standard language as in the case with Buddhist Hybrid Sanskrit can increase quality significantly. 

\begin{table}[t]
\begin{center}
\small
\begin{tabular}{lrrrrrr}
\toprule
& \multicolumn{3}{c}{\bf DCS lemmas} & \multicolumn{3}{c}{\bf MITRA-lexicon headwords, $n \geq 20$} \\
\cmidrule(lr){2-4}\cmidrule(lr){5-7}
\bf Reference & $\geq$1000 & 20--999 & verbs & sa--en & sa--bo & sa--zh \\
\midrule
Monier-Williams 1872 & 95.7 & 83.7 & 64.2 & 86.8 & 74.0 & 75.6 \\
Apte 1890 & 93.6 & 63.4 & 54.3 & 71.3 & 61.4 & 63.6 \\
MW $\cup$ Apte & 96.4 & 86.2 & 65.4 & 89.1 & 76.0 & 77.3 \\
Edgerton BHS & 32.7 & 12.6 & 1.0 & 16.2 & 27.3 & 32.7 \\
MW $\cup$ Apte $\cup$ BHS & 96.9 & 86.8 & 65.5 & 89.7 & 82.7 & 84.6 \\
Negi (inverted) & 78.9 & 53.7 & 3.5 & 62.5 & 70.8 & 71.3 \\
Hirakawa (inverted) & 92.0 & 46.0 & 32.1 & 57.7 & 75.8 & 83.2 \\
\midrule
Lemmas / headwords & 866 & 13{,}262 & 2{,}784 & 10{,}166 & 9{,}465 & 6{,}019 \\
\bottomrule
\end{tabular}
\end{center}
\caption{\label{tab:lemmacov} Lemma coverage of the reference dictionaries (in \%): the share of DCS lemmas, by number of occurrences in the DCS corpus, and of the single-word headwords of MITRA-lexicon with at least 20 attestations, that have an entry in the reference after normalization to DCS-style stems. Headwords are restricted to lemmas produced by the analyzer, i.e.\ without the sandhied surface entries of compounds.}
\end{table}

\subsection{Baselines: Statistical Aligners and the Teacher}
\label{sec:baselines}
\label{sec:baselines-refdict}
\label{sec:ablation}

In order to measure the technical performance of the aligner against other established word alignment techniques, we compare our trained student model with fast\_align \cite{dyer2013fastalign}, GIZA++ \cite{och2003systematic} (IBM Model~4, via mgiza) and eflomal \cite{ostling2016eflomal}, running them on the same corpora with Sanskrit lemmas as source tokens and Tibetan syllables, Chinese characters and English words as target tokens, and in both directions. As is standard, the two directional alignments are combined with the grow-diag-final-and heuristic \cite{koehn2003statistical}, which keeps the links found in both directions and adds links found in only one of them if they neighbor a kept link or connect two words that are still unaligned.
Table~\ref{tab:baselines} gives link-level $F_1$ against the held-out teacher labels of Section~\ref{sec:model}.
This measures agreement with the teacher rather than correctness, and it penalizes the baselines for linking function words that the teacher leaves untagged by design.\footnote{Restricted to the source words the teacher did tag, the precision of the statistical aligners rises to 53--75\%, against 95--98\% for the student.}
The statistical aligners agree with the teacher only about half of the time ($F_1$ of 42--63, against 86--89 for the student), in line with \newcite{jagadeeshan2026padamitra}, where fast\_align reaches a key Jaccard of 0.14 in a Sanskrit--English glossary task, against 0.52--0.72 for fine-tuned models (0.62--0.72 for instruction-tuned LLMs).
Students trained with the same data and recipe on other base models (middle of Table~\ref{tab:baselines}) show that the MITRA pretraining accounts for 4--7 points (81.1--83.6 off the shelf, against 85.9--88.0).
Gemma-2-MITRA lies in between (82.5--85.0), and a MITRA-pretrained 2B model stays 8--9 points behind the 9B model, and still 5--7 points when trained for three epochs, the same wall time.
Three epochs add only 1--2 points to the 9B model.
Against the expert gold alignments (last column), the students fall in the same order, except that Gemma-2-MITRA and the 9B model without Buddhist pretraining are on par (79.4 and 79.9), and the 9B model trained for three epochs is no better than the one trained for one (83.9 against 84.6).
All students outperform the statistical aligners by a wide margin.

\begin{table}[t]
\begin{center}
\small
\begin{tabular}{lrrrr}
\toprule
& \multicolumn{3}{c}{\bf Link $F_1$ vs.\ teacher labels} & \bf Expert gold \\
\cmidrule(lr){2-4}\cmidrule(lr){5-5}
\bf Aligner & sa--bo & sa--zh & sa--en & sa--en \\
\midrule
fast\_align & 51.7 & 54.4 & 42.0 & 48.3 \\
GIZA++ & 62.2 & 59.2 & 41.8 & 49.6 \\
eflomal & 62.8 & 61.6 & 46.6 & 53.9 \\
\midrule
\multicolumn{5}{l}{\textit{Students fine-tuned on the same teacher labels}} \\
Qwen3.5-2B, MITRA pretraining & 79.3 & 79.5 & 76.9 & 74.9 \\
\quad same, 3 epochs (compute-matched) & 82.5 & 82.7 & 79.1 & 75.5 \\
Qwen3.5-9B, no Buddhist pretraining & 81.5 & 83.6 & 81.1 & 79.9 \\
Gemma-2-MITRA 9B & 83.4 & 85.0 & 82.5 & 79.4 \\
MITRA Qwen3.5-9B-Base stage 2 & 88.0 & 87.6 & 85.9$^\dagger$ & \bf 84.6$^\dagger$ \\
\quad same, 3 epochs & \bf 88.9$^\dagger$ & \bf 89.2$^\dagger$ & \bf 86.9 & 83.9 \\
\bottomrule
\end{tabular}
\end{center}
\caption{\label{tab:baselines} Statistical aligners and students with different base models: link-level $F_1$ against 1{,}489 held-out sentence pairs labeled by the teacher (Section~\ref{sec:model}). The students are trained on the same five-direction data with the same recipe (one epoch unless stated otherwise, full fine-tuning, learning rate $2\cdot10^{-5}$) and differ only in the base model. $^\dagger$ marks the checkpoint that aligned the respective corpus (Section~\ref{sec:model}). Last column: link $F_1$ against the expert gold alignments of Section~\ref{sec:gold}.}
\end{table}

Built into lexica by the same lemma-to-sense counting, without the phrase entries and filters of the released resource, the statistical aligners recover the dominant equivalent of frequent headwords reasonably well (Table~\ref{tab:baselines-refdict}), for English almost as well as the student (62.4 for eflomal against 63.9), less so for Tibetan (62.6 against 77.5) and Chinese (55.0 against 71.1).
For headwords with at most four attestations, which make up most of the resource, their top-ranked sense is confirmed for only 5--14\% of the headwords, against 23--45\% for the student.

\begin{table}[t]
\begin{center}
\small
\begin{tabular}{lrrrrrr}
\toprule
& \multicolumn{3}{c}{\bf P@1, $n \geq 20$} & \multicolumn{3}{c}{\bf P@1, $n \leq 4$} \\
\cmidrule(lr){2-4}\cmidrule(lr){5-7}
\bf Aligner & sa--bo & sa--zh & sa--en & sa--bo & sa--zh & sa--en \\
\midrule
fast\_align & 53.0 & 51.4 & 59.5 & 9.1 & 7.5 & 11.4 \\
GIZA++ & 58.0 & 49.0 & 57.2 & 8.7 & 4.8 & 4.8 \\
eflomal & 62.6 & 55.0 & 62.4 & 7.1 & 7.1 & 14.3 \\
Student (this paper) & \bf 77.5 & \bf 71.1 & \bf 63.9 & \bf 45.1 & \bf 39.1 & \bf 22.8 \\
\bottomrule
\end{tabular}
\end{center}
\caption{\label{tab:baselines-refdict} Strict P@1 against Negi, Hirakawa and Monier-Williams~$\cup$~Apte (Section~\ref{sec:refdict}) for lexica built from each aligner's links, pooled over headwords with at least 20 and at most 4 attestations ($n$).}
\end{table}

The lexica could also have been mined with the teacher alone.
A second run of the teacher with identical settings on 155 of the held-out pairs (Table~\ref{tab:ablation}) shows that it does not fully agree with itself (span $F_1$ of 88.1--97.0, link $F_1$ of 94.5--96.6), and the student stays 3.5--7.5 points below this ceiling at span level and 3.3--7.6 at link level. Most of the disagreement does not concern the equivalent itself.
Of the 2{,}212 Sanskrit words on which student and teacher differ in all 1{,}489 held-out pairs, 70\% are tagged by only one of the two and 27\% receive overlapping spans of different extent (\textit{lus kyis} or \textit{lus} for \sa{k\=ayena}, {\zh 光明} or {\zh 光} for \sa{arci\d{s}a\d{h}}), and only 2.8\% are linked to different parts of the translation.
The two runs of the teacher differ in the same two ways (69\% and 31\%), so that the remaining disagreement reflects the latitude of the labeling policy rather than wrong equivalents.
The teacher alone would, however, have cost about USD~45{,}000 for the 1.30 million sentence pairs at the list prices of October 2026 (USD~0.026--0.041 per pair, mostly for 6{,}700--10{,}600 reasoning tokens, and half of that through the batch interface), whereas the teacher--student setup called the teacher 12{,}251 times and ran the student for about 36 GPU-hours (Section~\ref{sec:inference}).

\begin{table}[t]
\begin{center}
\small
\begin{tabular}{lrrrrr}
\toprule
& & \multicolumn{2}{c}{\bf Span $F_1$} & \multicolumn{2}{c}{\bf Link $F_1$} \\
\cmidrule(lr){3-4}\cmidrule(lr){5-6}
\bf Pair & \bf Sent.\ pairs & \bf Teacher & \bf Student & \bf Teacher & \bf Student \\
\midrule
Sanskrit--Tibetan & 47 & 88.1 & 83.5 & 94.5 & 90.9 \\
Sanskrit--Chinese & 65 & 93.8 & 90.3 & 96.6 & 93.3 \\
Sanskrit--English & 43 & 97.0 & 89.5 & 96.6 & 89.0 \\
\bottomrule
\end{tabular}
\end{center}
\caption{\label{tab:ablation} Agreement with the held-out teacher labels on a subset of 155 sentence pairs: a second run of the teacher under identical settings, and the student (the checkpoint that aligned the respective corpus), both scored on the same sentences. Span $F_1$ as in Section~\ref{sec:model}, link $F_1$ as in Table~\ref{tab:baselines}.}
\end{table}

\subsection{Gold Alignments for Sanskrit--English}
\label{sec:gold}

We drew 100 sentences of 6--21 analyzed words from the Sanskrit--English alignment output (38 texts), excluding texts transmitted with resolved sandhi so that the analyzer is evaluated as well.
Three Sanskritists who hold PhD degrees in Indology and who extensive experience in reading and annotating Sanskrit resources, correct the model's prediction in a dedicated interface that shows the Sanskrit, the translation, the analyzed words and the entries of Monier-Williams, Apte, Edgerton and Böhtlingk--Roth. All three annotate every sentence, so that their agreement can be measured.
Each analyzed word is one unit, and so is each compound as a whole.
For each unit the annotator marks either the English span that renders it, or states that it has none.
As in the teacher prompt, the span is one contiguous stretch with only the word that carries the meaning, without articles, prepositions, auxiliaries or pronouns (\textit{go} in \textit{I go} for \sa{gacch\=ami}), except that a compound rendered as one chain is aligned as a whole (\textit{the departure of the lover} for \sa{priya-pray\=a\d{n}e}).
The three annotators (A, B and C) have annotated all 100 sentences, 1{,}444 units (1{,}246 words and 198 compounds).
Since the annotators correct the model's prediction rather than starting from scratch, both their agreement and the model's scores below are biased toward the model.

\paragraph{Inter-annotator agreement.}
Pairs of annotators make the same decision on 77.6--84.8\% of the units and agree on whether a unit has an English equivalent at all on 81.6--87.4\% ($\kappa$ of 0.60--0.73).
Their spans match exactly on 80.1--85.6 in terms of $F_1$, and where both align a unit their spans overlap almost completely (mean Jaccard 0.95--0.97).
Krippendorff's $\alpha$ over the three annotators, with the decision as a nominal value, is 0.77.
The disagreement lies mostly in whether a unit is aligned at all, not in the extent of the span (688 of the 854 differing decisions of the three pairs).
The annotators differ in how much they align (A 65\%, B 73\% and C 59\% of the units, against 57\% for the model).
In its pairs with A and with C, B aligns 419 units that the other leaves unaligned, against 87 the other way round, and 188 of these 506 are pronouns, particles, conjunctions or the copula (\sa{tad}, \sa{ca}, \sa{na}, \sa{idam}, \sa{eva}, \sa{yad}), which the guidelines exclude.
Without these function words $\alpha$ rises to 0.81.
Following \cite{krippendorff2004content}, values of $\alpha \geq 0.80$ are conventionally regarded as reliable, whereas values between 0.667 and 0.80 permit only tentative conclusions.

\paragraph{Model against the gold alignments.}
Table~\ref{tab:gold} compares the spans from which the lexicon was built with each annotator.
Precision is high and stable across annotators (89--91 under exact match), and recall varies with how much the annotator aligns (70--88).
Against A, B and C, $F_1$ is 85.6, 78.2 and 88.4, in the range of their agreement with each other (80.1--85.6).
Against the majority decision (the 1{,}404 units on which at least two annotators agree) it is 88.4, and on the 1{,}037 units on which all three agree 95.3.
Annotators left 74--88\% of the units unchanged.
Most corrections add a missing alignment (68\% over the three annotators, half for C, who also removes the most).
For compounds, recall is only 52--69\%, since a compound receives a span only when all of its members are aligned and their equivalents form one unit in the translation (Section~\ref{sec:build}). Precision stays at 92--94.
\begin{table}[t]
\begin{center}
\small
\begin{tabular}{lrrrrrr}
\toprule
\bf Gold & \bf Sent. & \bf Units & \bf P & \bf R & \bf $F_1$ & \bf Unch. \\
\midrule
Annotator A & 100 & 1{,}444 & 91.3 & 80.6 & 85.6 & 84.0 \\
Annotator B & 100 & 1{,}444 & 89.0 & 69.7 & 78.2 & 74.4 \\
Annotator C & 100 & 1{,}444 & 89.4 & 87.5 & 88.4 & 87.8 \\
Majority (2 of 3) & 100 & 1{,}404 & 92.3 & 84.8 & 88.4 & 87.0 \\
All three agree & 100 & 1{,}037 & 95.7 & 94.9 & 95.3 & 93.8 \\
\midrule
\multicolumn{7}{l}{\textit{Majority, by unit kind}} \\
\quad words & 100 & 1{,}220 & 92.0 & 88.9 & 90.4 & 89.3 \\
\quad compounds & 100 & 184 & 95.6 & 62.3 & 75.4 & 71.2 \\
\bottomrule
\end{tabular}
\end{center}
\caption{\label{tab:gold} Sanskrit--English alignments of the model against the expert gold alignments of the 100 sentences of the agreement study: precision, recall and $F_1$ of the spans under exact match, and the share of units the annotator left unchanged. ``Majority'' takes the decision of at least two of the three annotators (units without one are left out), ``all three agree'' only the units on which the three annotations are identical.}
\end{table}

\section{Conclusion}
\label{sec:conclusion}

We have presented MITRA-lexicon, lexica with Sanskrit as source and English, Tibetan and Chinese as targets, in which every sense is a count over a corpus and every count can be traced to its sentences: 2.18 million entries with 10.4 million attestations, to our knowledge the first resource of this kind for these language pairs.
Against the reference dictionaries of Monier-Williams and Apte, Negi and Hirakawa, the top-ranked sense of headwords with at least 100 attestations is confirmed in 75--90\% of the cases and one of the top five in 94--99\%.
Agreement is lower for rare headwords, but these figures are lower bounds, as of 60 reviewed headwords attested only once, none was misaligned, 39 had a correct and 15 a nearly correct equivalent, and all remaining errors lay in the Sanskrit key or lemma rather than in the alignment.
The resource also reaches beyond the references, since about a quarter of the frequent headwords of the Sanskrit--Tibetan and Sanskrit--Chinese lexica have no entry in Monier-Williams or Apte.
On expert-corrected Sanskrit--English alignments, the spans of the model reach a precision of about 90\% and agree with each annotator about as closely as the annotators agree with each other.
The lexica were built with existing tools, a Sanskrit morphological analyzer, a language model adapted to the languages and 9{,}366 examples labeled by a commercial model, and the resulting aligner agrees with these labels far more closely than statistical aligners do ($F_1$ of 86--89 against 42--63).
Everything needed to rebuild or extend the lexica is released.
In future work we plan to normalize the target side so that inflectional variants of one equivalent are merged, to add P\=ali and further Sanskrit--English translations as they become available, and to extend the annotations of the DCS to include non-classical and non-vedic language, which is especially important for covering the specific language of Buddhist texts.

\section{Limitations}
\label{sec:limitations}

Gold alignments exist only for Sanskrit--English, so the evaluation is uneven. English is also the target language for which the underlying language model is strongest and the only one with explicit word boundaries, so the alignment quality measured there may not carry over to Tibetan and Chinese. For those pairs the only evidence is agreement with reference dictionaries, which evaluates entries rather than alignments and gives a lower bound (Section~\ref{sec:refdict}).
The English lexicon is clearly dominated by a few large texts, as the epics and pur\=a\d{n}as account for about half of the Sanskrit--English sentence pairs.
Because glosses are recorded as they occur, the evidence for one equivalent can be split over senses that differ only in case particles or inflection (for example \textit{snying rje}, \textit{snying rje'i}, \textit{snying rjes} for \sa{karu\d{n}\=a}). Merging them requires lemmatization on the target side in case of Tibetan, which we leave to future work.
Errors of the ByT5-Sanskrit analyzer and of the Sanskrit e-texts propagate into entries under wrong lemmas or with wrong compound boundaries.
Our filter for them (Section~\ref{sec:build}) addresses only typing errors one edit away from a frequent word and has been inspected only cursorily, and the noise in the e-texts and in their analysis would require more dedicated attention.
Finally, the training signal comes from a commercial model.
The labeled seeds are released, so that the student can be reproduced from them, but reproducing the labeling stage itself requires access to that model and might therefore become impossible in this specific configuration in the future.

\section*{Acknowledgements} We thank the Tsadra Foundation for their generous support in providing funding for the computational infrastructure that made this study possible. We also acknolwedge the use of generative AI (Claude Code) in assisting with the programming needed for the study, and with the language of the paper.

\bibliographystyle{acl}
\bibliography{main}
\end{document}